\documentclass[letterpaper, 10 pt, conference]{ieeeconf}
\IEEEoverridecommandlockouts  
\PassOptionsToPackage{hidelinks}{hyperref}
\usepackage{cite}
\usepackage{amsmath,amssymb,amsfonts}
\usepackage{graphicx}
\usepackage{booktabs}
\usepackage{textcomp}
\usepackage{xcolor}
\usepackage{url}
\usepackage{tikz}
\usetikzlibrary{arrows.meta,positioning,fit}
\IfFileExists{orcidlink.sty}{\usepackage{orcidlink}}{\newcommand{\orcidlink}[1]{}}

\begin{document}

\title{\LARGE \bf
Borrowed Eyes: Markerless Nano-UAV Flight with an Active Quadruped Observer
}

\author{Alejandro Lorite Mora$^{1}$, Dimitrios Arapis$^{2}$ and Andr\'es Fa\'i\~na$^{1}$%
\thanks{This work has been submitted to the IEEE for possible publication. Copyright may be transferred without notice, after which this version may no longer be accessible.}%
\thanks{This research was supported by the IT University of Copenhagen and Helix Lab in Denmark under the Novo Nordisk Fonden through project grant number 224611. We also thank Rasmus Finderup Thomsen for his help at the Helix Lab Robotics Lab.}%
\thanks{$^{1}$IT University of Copenhagen, Denmark {\tt\small \{lori, anfv\}@itu.dk}. A.~Lorite Mora is also with Helix Lab and Novo Nordisk A/S, Denmark.}%
\thanks{$^{2}$Novo Nordisk A/S, Hiller\o d, Denmark\newline {\tt\small dim.arapis@outlook.com}}%
}

\maketitle
\thispagestyle{empty}
\pagestyle{empty}

\begin{abstract}
Nano unmanned aerial vehicles (nano-UAVs) can navigate confined spaces that larger robots cannot, but their payload capacity severely limits the sensors and compute available for self-localization in global navigation satellite system (GNSS)-denied environments. We present a vision-based system that localizes a nano-UAV from a quadruped robot with an arm-mounted camera. The quadruped tracks the drone, estimates its position in its own coordinate system using segmentation masks and depth, and transmits that position over a real-time radio link. To ensure continuous tracking, we developed a perception-aware nonlinear Model Predictive Controller (NMPC) that dynamically adjusts the quadruped's body and arm to maximize the drone's visibility, treating observation reliability as a primary control objective. Relying solely on this external estimate, the nano-UAV executed predefined trajectories from takeoff to landing with a median 3D localization error of 59\,mm. The system demonstrates robustness against short visual occlusions by smoothly transitioning to onboard inertial flight when line-of-sight is temporarily lost. Ultimately, this framework allows a quadruped to offload the localization burden of a nano-UAV, enabling inspection of complex spaces that neither robot could navigate alone.
\end{abstract}

\section{Introduction}

Industrial assets such as storage tanks, pressure vessels, pipes, and tunnels require regular inspection for corrosion, cracks, and other structural damage. Manual inspection is time-consuming, costly, and hazardous, especially in narrow or elevated spaces. Nano unmanned aerial vehicles (nano-UAVs), weighing only tens of grams, are well suited to reaching and inspecting areas inaccessible to larger robots.

\begin{figure}[!t]
\centering
\vspace*{5.8pt}%
\includegraphics[width=0.8\columnwidth]{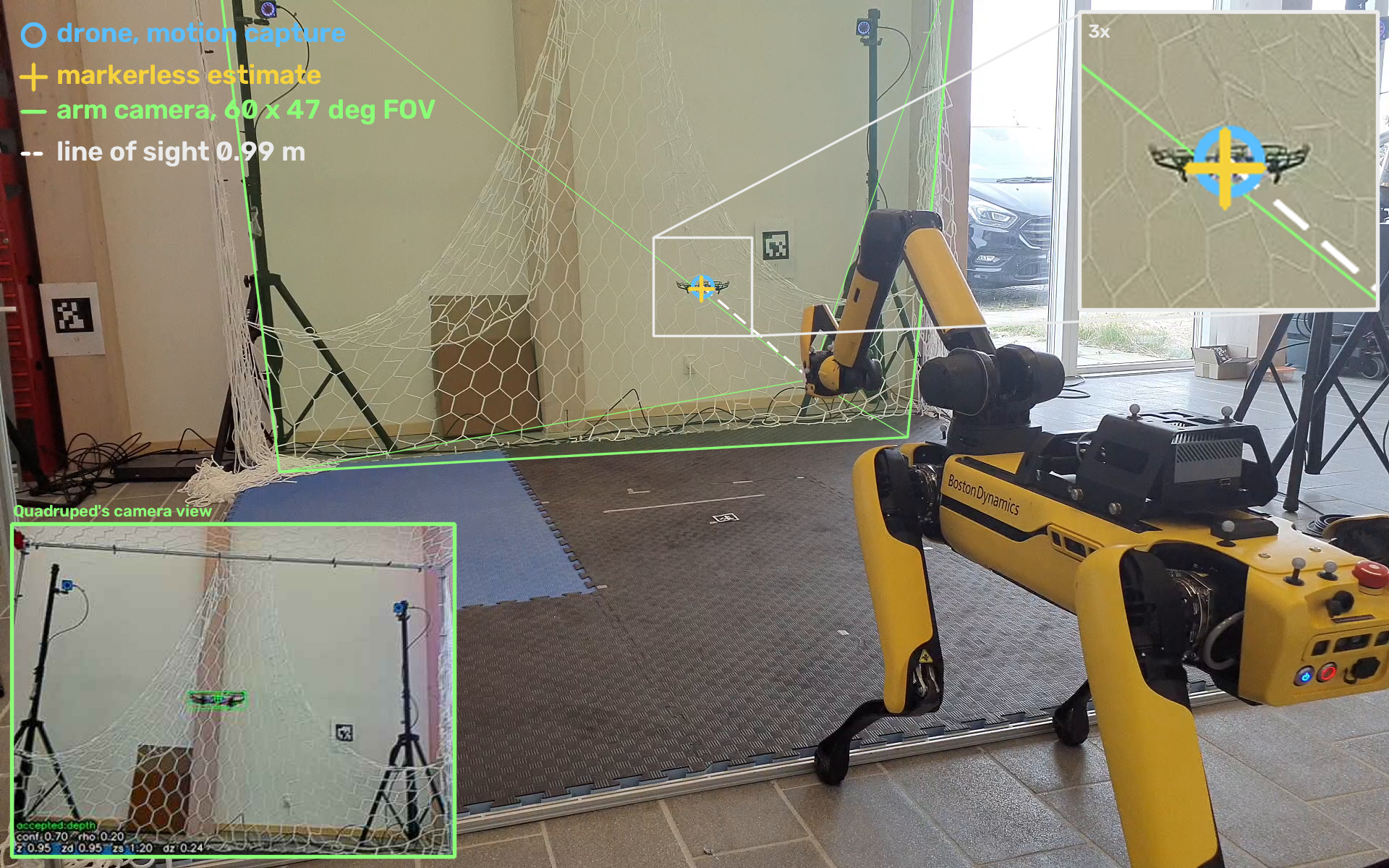}
\caption{The quadruped localizes the nano-UAV with no marker on the drone and no localization sensing aboard it. It computes the drone's position with its arm camera, in its own map, and sends it over radio. Overlays are projected into the footage from the trial's own recording: the yellow cross is the markerless estimate the drone flew on, the blue circle is the motion-capture ground truth used for evaluation only. Insets: the quadruped's camera view and a $3\times$ magnification of the drone.}
\label{fig:system}
\end{figure}

However, autonomous nano-UAV deployment is severely constrained by onboard localization. Larger UAVs can carry stereo cameras or LiDAR to run visual-inertial odometry (VIO) or simultaneous localization and mapping (SLAM), but these sensors exceed the payload and power capacity of nano-scale platforms~\cite{niculescu_nanoslam_2024}. Confined industrial settings also feature repetitive, textureless steel surfaces and uneven illumination, which degrade feature-based localization even when sufficient computation is available.

In laboratory settings, nano-UAVs rely on external localization infrastructure like motion capture or ultra-wideband (UWB) beacons. However, installing fixed infrastructure across industrial facilities is costly, inflexible, and often impractical. Operating in real environments requires localization that preserves the nano-UAV's payload margin without permanent infrastructure.

Rather than increasing the sensing burden on the drone or instrumenting the environment, we offload localization to a mobile ground robot with greater sensing and computational capacity (Fig.~\ref{fig:system}). The quadruped localizes itself and estimates the 3D position of the nano-UAV using an arm-mounted camera, while the drone flies using this externally supplied estimate. Crucially, the estimate is expressed directly in the quadruped's global coordinate map rather than in an onboard frame. This provides a shared spatial representation for heterogeneous collaboration, allowing the nano-UAV to scout confined spaces and report points of interest in a frame the ground robot can directly use.

Our prior work~\cite{loritemora_clawar_2026} demonstrated this cooperative division of labor using fiducial markers on a nano-UAV~\cite{olson_apriltag_2011}. However, physical markers add considerable payload mass, increase aerodynamic drag, and enlarge the vehicle's footprint. In addition, reactive camera tracking often loses visual contact during fast maneuvers or temporary occlusions.

To overcome these limitations, we introduce a markerless localization and active observation framework for nano-UAVs. Our main contributions are: (i)~\textbf{Markerless external localization:} A position-only perception pipeline that estimates the 3D position of a bare nano-UAV in the quadruped's map frame, using depth-integrity gates to discard corrupted measurements and convert sensor failures into safe dropouts. (ii)~\textbf{Perception-aware active observation:} A nonlinear Model Predictive Controller (NMPC) that coordinates the quadruped's body and arm motion to preserve line-of-sight and optimize viewing geometry. (iii)~\textbf{Closed-loop flight under intermittent perception:} Closed-loop flight validation from takeoff to landing on the external vision feed alone, demonstrating autonomous recovery from visual occlusions through onboard inertial coasting.

\section{Related Work}
\label{sec:rw-markerless}

\textbf{Ground-Aerial Collaborative Localization.} Leveraging ground robots to assist UAV localization is an established strategy for global navigation satellite system (GNSS)-denied environments. Existing approaches utilize ultra-wideband radios on both platforms to achieve decimeter-level localization~\cite{queralta_uwb_2020}, employ ground vehicles as external references to refine onboard state estimates~\cite{depetrillo_ugv_uav_2024}, or use legged robots as mobile magnetic beacons, achieving tracking errors of 8 to 11\,cm~\cite{brunacci_flytrackland_2026}. These methods either require dedicated onboard payload hardware~\cite{depetrillo_ugv_uav_2024} or provide lower accuracy than close-range visual localization~\cite{queralta_uwb_2020,brunacci_flytrackland_2026}.

Markerless inter-robot observation has been explored in multi-robot SLAM, where mutually observing robots use visual detections to refine their state estimates~\cite{rueggeberg_markerless_2026}. However, such observations have not been demonstrated as the primary external position source for closed-loop flight of a nano-UAV. Removing fiducial markers eliminates added mass and aerodynamic penalties, but requires localizing a small, dark, weakly textured, and symmetric airframe directly from appearance and depth. While render-and-compare 6D pose estimators~\cite{wen_foundationpose_2024,leeAny6DModelfree6D2025} address similar challenges, we exploit the flight controller's internal attitude estimation and compute only 3D position, reducing the vision pipeline to the essential external state (Section~\ref{sec:posestimator}).

\textbf{Vision-Based Localization of Small Aerial Targets.} Prior work on aerial target detection predominantly targets larger platforms. Vrba et al.~\cite{vrba_noncoop_2019} detected a 0.55\,m quadrotor using stereo vision up to 50\,m, with subsequent work reporting ranges up to 20\,m~\cite{vrba_mav_2020}. Carrio et al.~\cite{carrio_depthmaps_2020} localized drones using onboard depth maps at 16\,Hz up to 8\,m, but evaluated a 517\,mm platform and attributed range limits to sensor constraints rather than target dimensions. In LiDAR-based studies~\cite{vrba_lidar_2025}, classification degrades when a target returns fewer than 40 points. A bare nano-UAV challenges these limits at much closer range, where its small surface area returns very few depth points. We explicitly quantify this relationship up to 3.1\,m in Section~\ref{sec:exp-perception}.

\textbf{Aerial Navigation Through Narrow Apertures.} Navigating confined spaces has been addressed through collision-tolerant platforms operating through manhole-sized openings~\cite{depetris_manhole_2020,tranzatto_cerberus_2022}, or commercial caged inspection drones with full onboard SLAM.\footnote{Flyability Elios 3, \url{https://www.flyability.com/elios-3/}} Both strategies require structural protection and compute payloads well beyond the capacity of nano-UAVs. This leaves an important capability gap for nano-UAVs, which can access tighter spaces but cannot carry the heavy sensor packages assumed by existing platforms.

\section{Method}
\label{sec:method}

\subsection{System overview}
\label{sec:overview}

The proposed system is shown in Fig.~\ref{fig:system} and its processing pipeline in Fig.~\ref{fig:pipeline}. A Boston Dynamics Spot quadruped serves as the mobile base, equipped with its own arm-mounted gripper camera, which provides color and registered time-of-flight depth (Section~\ref{sec:camera}), and an NVIDIA Jetson AGX Orin onboard computer. This computer handles quadruped control, arm articulation, and drone pose estimation. From each RGB-D frame, the pipeline extracts at most one 3D position of the drone, referenced to the quadruped's map frame. Validated position estimates are transmitted via a low-latency radio link to the drone (a Bitcraze Crazyflie relying solely on an inertial measurement unit (IMU)) using the \texttt{crazyswarm2} server.\footnote{Crazyswarm2, \url{https://github.com/IMRCLab/crazyswarm2}}

The quadruped tracks its motion via onboard odometry, anchored to the world map frame by a static transform ($T_{\text{map}\to\text{odom}}$ in Eq.~\eqref{eq:chain}). In these experiments, an OptiTrack motion-capture system supplies that anchor and records ground truth for calibration and evaluation. Motion capture never provides state estimates to the flight controller: the drone completes its entire flight profile, from takeoff to landing, relying on the markerless pipeline. Anchoring instead to the quadruped's onboard SLAM map adds negligible error: Spot's internal map drifted $<1$\,cm (P95). We use the motion-capture anchor to keep reported tracking performance independent of ground-robot self-localization error.

\subsection{Mask-Based Position Estimation}
\label{sec:posestimator}
The drone's map-frame position is constructed from rigid transforms anchored to the quadruped's self-localization, closed by a markerless camera-to-drone estimate:
\begin{equation}
\mathbf{p}_{\text{map}\to\text{drone}}
 = T_{\text{map}\to\text{odom}}\!
   T_{\text{odom}\to\text{body}}\!
   T_{\text{body}\to\text{cam}}\!
   \underbrace{\mathbf{p}_{\text{cam}\to\text{drone}}}_{\text{estimated here}}
\label{eq:chain}
\end{equation}
Here $T_{A\to B}$ is the pose of frame $B$ expressed in frame $A$, so each factor carries a point from $B$ into $A$, acting on the homogeneous point. The first three encapsulate the quadruped's state: its localization in the map frame, its internal odometry, and the arm's forward kinematics. Only the final link is derived visually. Because the drone's IMU maintains its attitude and yaw, full 6D pose estimation is unnecessary. The visual problem reduces to estimating the drone's 3D position in the camera frame, extracted from a 2D segmentation mask $m_\phi(I)$ and the registered depth map $D$, where $D[\cdot]$ reads the finite, positive depth returns under a mask:
\begin{equation}
\begin{split}
(u,v) &= \operatorname{centroid}\big(m_\phi(I)\big), \\
z &= \operatorname{agg}\big(D[m_\phi(I)]\big) + \delta, \\
\mathbf{p}_{\text{cam}\to\text{drone}} &= z\,K^{-1}[u,\;v,\;1]^{\top},
\end{split}
\label{eq:posonly}
\end{equation}
enabling 3D unprojection via camera intrinsics $K$. Because depth returns come from the near face, a fixed offset $\delta=4.5$\,cm carries the reading to its center.

\subsubsection{Sensing and Detection}
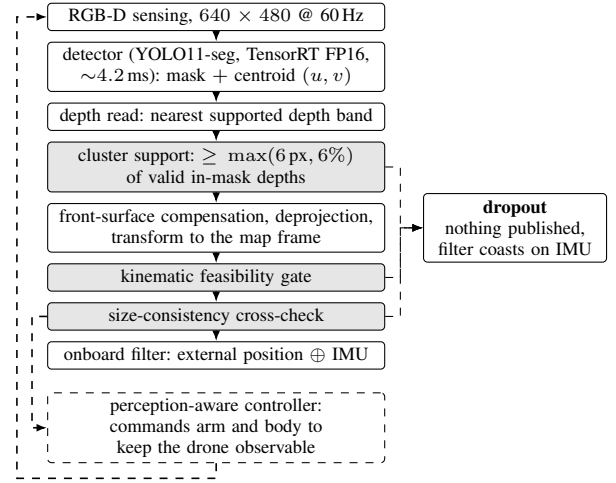
\begin{figure}[!t]
\centering
\begin{tikzpicture}[
    font=\scriptsize,
    node distance=1.4mm,
    stage/.style={draw, rounded corners=1.5pt, align=center, inner sep=2pt,
                  text width=0.50\columnwidth},
    gate/.style={stage, fill=black!10},
    drop/.style={draw, rounded corners=1.5pt, align=center, inner sep=2.5pt,
                 text width=0.26\columnwidth, fill=white},
    arr/.style={-{Latex[length=1.4mm]}, semithick},
    rej/.style={-{Latex[length=1.4mm]}, dashed},
    loop/.style={draw, dashed, rounded corners=1.5pt, align=center, inner sep=2pt,
                 text width=0.50\columnwidth},
    fb/.style={-{Latex[length=1.4mm]}, semithick, dashed},
]
\node[stage] (sensor) {RGB-D sensing, $640\times480$ @ 60\,Hz};
\node[stage, below=of sensor] (det) {detector (YOLO11-seg, TensorRT FP16, ${\sim}4.2$\,ms): mask $+$ centroid $(u,v)$};
\node[stage, below=of det] (depth) {depth read: nearest supported depth band};
\node[gate, below=of depth] (sup) {cluster support: $\geq \max(6\text{\,px}, 6\%)$ of valid in-mask depths};
\node[stage, below=of sup] (metric) {front-surface compensation, deprojection, transform to the map frame};
\node[gate, below=of metric] (kin) {kinematic feasibility gate};
\node[gate, below=of kin] (size) {size-consistency cross-check};
\node[stage, below=of size] (ekf) {onboard filter: external position $\oplus$ IMU};
\node[drop, right=5mm of metric.east, anchor=west] (drop)
      {\textbf{dropout}\\ nothing published,\\ filter coasts on IMU};
\node[loop, below=3mm of ekf] (obs)
      {perception-aware controller: commands arm and body to keep the drone observable};
\foreach \a/\b in {sensor/det, det/depth, depth/sup, sup/metric, metric/kin, kin/size, size/ekf}
    \draw[arr] (\a) -- (\b);
\foreach \g in {sup, kin, size}
    \draw[rej] (\g.east) -- ++(2mm,0) |- (drop.west);
\draw[fb] (size.west) -- ([xshift=-2mm]size.west) |- (obs.west);
\draw[fb] (obs.south) -- ++(0,-2mm) -| ([xshift=-4mm]sensor.west) -- (sensor.west);
\end{tikzpicture}
\caption{The deployed system. Solid path: one position per frame, expressed in the quadruped's own map frame. Shaded stages are the three integrity gates of the deployed gripper-camera path. A failed gate publishes nothing and the drone's filter coasts on its IMU, a dropout instead of a wrong position. Section~\ref{sec:exp-perception} reports their aggregate effect. Dashed: the accepted position also drives the perception-aware controller (Section~\ref{sec:pac}), which commands the arm and body so the drone stays observable.}
\label{fig:pipeline}
\end{figure}

For a nano-UAV of this scale, reliable depth extraction poses a greater challenge than 2D masking. Fig.~\ref{fig:pipeline} outlines the deployed pipeline. The gripper camera streams color and time-of-flight depth at $640\times480$ and $60$\,Hz, with the depth registered to the color frame on the robot itself. A YOLO11 instance-segmentation model,\footnote{Ultralytics YOLO11, \url{https://github.com/ultralytics/ultralytics}} exported to TensorRT at FP16, segments the drone in each color frame in about $4.2$\,ms on the onboard computer.

\subsubsection{Auto-Labeled Training Data}
To bypass manual annotation, we trained the segmentation model on in-domain imagery from an automated pipeline. We prompted a foundation model (SAM 2~\cite{ravi_sam2_2024}) using projected motion-capture positions of the drone. Motion capture is used solely as a prompting mechanism for the labeling pipeline: the live detector never sees external motion-capture data, ensuring no circularity in training. The training set holds $7\,306$ positive frames: $6\,025$ auto-labeled from nine recordings of the drone hanging and in flight, and $1\,281$ composites that paste segmented drones onto footage of the surrounding rooms recorded without the drone, so the detector meets the route's backgrounds before it flies there. A further $1\,099$ drone-free frames supply negatives, and two recordings serve as test sets. Seven retraining runs as the dataset expanded required $4.2$ GPU-hours across laptop and desktop GPUs, with the deployed model fine-tuned in $0.31$\,h with zero manual labels.

\subsubsection{Depth Extraction and Integrity Gates}
The range aggregation $z$ is sensor-dependent. While stereo cameras allow median depth extraction, the time-of-flight gripper camera requires selecting the nearest supported depth band, since the drone returns only $6$ to $14\%$ of the valid depths inside its own mask and the median therefore locks onto the background. By construction, the drone is the nearest surface inside a correct mask. Replayed against motion-capture truth, with the aggregation isolated from detector quality, the nearest band places the drone within $250$\,mm in $93$ to $96\%$ of frames with a $20$\,mm median error, versus $12$ to $17\%$ at $1.8$ to $2.4$\,m for the median.
 Crucially, thin airframes severely degrade depth readings. To prevent the estimator from silently locking onto the background, we gate every range before publishing. On the gripper camera, three gates carry it:
\begin{enumerate}
    \item \emph{Cluster support}: a band must hold at least $\max(6\text{\,px}, 6\%)$ of the valid in-mask depths, or no range is produced,
    \item \emph{Kinematic feasibility} (rejecting updates implying $>0.8$\,m/s jumps),
    \item \emph{Size consistency} against expected projection size.
\end{enumerate}
Our earlier stereo pipeline adds a depth-completeness floor ($\rho \geq 0.40$) and a background-annulus rejection. The nearest-band search makes both redundant on the time-of-flight camera: a low valid fraction is the regime that search exists to rescue, and a mask holding only background yields no band in front of it.
If any gate fails, nothing is published, and the drone coasts on its IMU. The relay forwards valid positions without extrapolation and commands landing if outages exceed safety limits.

\subsection{Camera, Calibration, and Ground Truth}
\label{sec:camera}
\label{sec:calib}
The gripper camera's pose is intrinsically derived from the quadruped's kinematic chain. Its time-of-flight depth successfully returns ranges on the dark airframe where a stereo camera (e.g., Intel RealSense D435) fails. Accuracy metrics compare the estimated drone position against motion-capture ground truth. For the gripper camera, the tracking chain relies on the quadruped's kinematics registered to the motion-capture frame. We report raw accuracy through the calibrated chain alongside precision after rigid Horn/Markley alignment of the estimated and reference tracks, which absorbs a constant offset in the reference chain and so isolates the estimator's intrinsic precision. Experiments evaluate three configurations: the deployed bare airframe, an airframe with a ${\sim}1.8$\,g matte panel restoring stereo depth (diagnostic), and an airframe equipped with an AprilTag (fiducial baseline).

\subsection{Single-Estimator Fusion, IMU Coasting, and Latency}
\label{sec:fusion}
\label{sec:coast}
The drone operates an extended Kalman filter (EKF) that fuses the markerless position updates with high-rate attitude and yaw estimates from its onboard IMU (Bosch BMI088). The drone maintains no map frame of its own. Between external position updates, the filter integrates IMU measurements to coast, with uncertainty swelling during coasting and shrinking upon correction. Visual processing and radio transmission introduce a pipeline latency of ${\sim}180$\,ms, so each estimate describes where the drone was, not where it is. The flights reported here fuse every estimate on arrival, without latency compensation, which leaves a position lag along the velocity vector. The measurement noise of each estimate is set to 5\,cm, matching the measured error of the markerless pipeline, rather than to the millimeter trust used for motion capture. The residual lag is one source of the tracking error in Section~\ref{sec:exp-tracking}.

\subsection{Perception-Aware Active Observation}
\label{sec:pac}
To maintain continuous localization, the quadruped must actively keep the drone near the camera's optical center and within its reliable depth range. A naive controller aiming at the last known position lags behind fast maneuvers, causing the drone to exit the field of view. Instead, we formulate observation quality as the primary objective of a nonlinear model predictive controller (NMPC), commanding the quadruped's body and arm to anticipate the drone's motion using its known trajectory. Over a receding horizon of $N$ steps, the observer minimizes:
\begin{equation}
\begin{split}
J = \sum_{k=0}^{N-1} \Big(\; & w_c\,\theta_k^{2} + w_r (d_k - d^{\star})^{2} - w_e \log p_k \\
 &+ w_b\,\nu_k^{2} + \lVert \mathbf{u}_k \rVert_R^{2} + w_m s_k^{2} + \Omega_k \Big)
\end{split}
\label{eq:pac}
\end{equation}
Here, $\theta_k$ is the angular deviation from the optical axis (the dominant penalty, as reliability degrades beyond ${\sim}30^\circ$). $d_k$ is the distance to the drone, penalized against a standoff target $d^{\star}=1.0$\,m (flown median $1.25$\,m, Table~\ref{tab:control}). The estimator's valid-publish probability $p_k$ is pre-tabulated against range and off-axis angle from empirical limits (Section~\ref{sec:exp-perception}). $\nu_k$ penalizes image-plane velocity (motion blur), $\mathbf{u}_k=(\mathbf{u}_b,\mathbf{u}_h)$ are body and hand velocities regularized by $R$, and $s_k$ softens a 40\,px image-border margin constraint. $\Omega_k$ collects four smaller regularizers that keep the posture natural and the commands smooth: arm reach toward a comfortable mid-reach, body-arm alignment, a soft ceiling on hand height, and a penalty on the command change between steps. The last two were added after closed-loop trials drove the shoulder into its joint limits. We re-solve Eq.~\eqref{eq:pac} over an $N{=}8$ horizon at 5\,Hz and stream the planned commands at 10\,Hz. The matrix $R$ encodes that arm motion is cheap and rapid, whereas body motion is expensive. The reactive baseline we compare against in Section~\ref{sec:exp-tracking} is a visual servo with no prediction and no cost trade-off. At $2$\,Hz, it places the hand a fixed distance short of the last received drone position along the body-to-drone ray, aims the camera at that position, and walks the body only when the target leaves the arm's reach. Because the trajectory is uploaded pre-flight, the radio link transmits only position updates, meaning path-tracking accuracies cleanly reflect localization performance.

\subsection{Position-only control and the rate floor}
\label{sec:budget}
\label{sec:poc}
Two preliminary results set the estimator's design targets. First, we send the drone only its position, and its EKF takes attitude and yaw from the IMU, as ultra-wideband and Lighthouse positioning systems do~\cite{queralta_uwb_2020}. Flown on motion capture, this position-only mode matched full-pose control on every paired trajectory. Only yaw is given up: it drifts 9 to $10^\circ$ per minute, against $0.8^\circ$ under full-pose control, a bounded wander rather than a divergence. Second, we measured what the flight demands of the position feed. A shim degrades the motion-capture position one axis at a time, in update rate, latency, noise, and dropout, and the drone flies on the degraded feed. Only the rate axis resolved a clean crossing in this single-pass session: hover remained stable down to 5\,Hz and broke down at 1\,Hz, so the markerless pipeline must publish above 5\,Hz, and Section~\ref{sec:exp-perception} shows that it does, at about four times that rate. The latency, noise, and dropout axes remain preliminary.

\emph{Code and data.} All the software described in this paper, comprising the position estimator, the labeling tools, the drone firmware and the changes to crazyswarm2, together with the recorded trials, will be published on acceptance.

\section{Experiments}
\label{sec:experiments}

Three experimental campaigns evaluate the system against our contributions: (A) markerless localization accuracy and depth integrity gating, (B) active observation and predictive tracking, and (C) closed-loop flight with no external localization in the flight loop, including visual occlusions and envelope limits. Together, these experiments demonstrate that the quadruped tracks the bare nano-UAV, maintains visibility, and closes the flight control loop on visual estimates alone.

\subsection{Markerless localization accuracy and integrity gating}
\label{sec:exp-perception}

We evaluate localization accuracy against motion-capture ground truth under two regimes. Offline, on hovering flight recordings with a stationary arm ($1.6$ to $1.8$\,m distance), we evaluate static estimation accuracy across $205$ ground-truth synchronized frames. Live, we evaluate the system across eight closed-loop flights ($1\,069$\,s total airborne time) with an actively moving camera spanning $0.6$ to $2.3$\,m stand-off, measuring accuracy under real operational conditions.

\emph{Accuracy and coverage.} On the hovering recording, the detector identifies the drone in $89\%$ of frames, yielding a median 3D error of $38.6$\,mm ($32.8$\,mm lateral). In live flight (Fig.~\ref{fig:gripper}), the median 3D error across $17\,363$ positions is $49$\,mm (P95 $119$\,mm), with lateral and along-axis components contributing equally ($31$\,mm median each). Only $0.6\%$ of positions deviate by more than $300$\,mm from truth. Live errors are slightly higher than static figures because the camera moves dynamically. To isolate camera motion from drone motion, we conducted trials tracking a hovering drone: over a shared distance band, a stationary camera achieves $43$\,mm median error versus $55$ to $58$\,mm when tracking, showing that camera motion incurs only a modest accuracy penalty. The pipeline publishes updates at $20$\,Hz whenever the drone is visible, comfortably exceeding the $5$\,Hz flight stability threshold. In an unmapped room absent from training data, background clutter caused false detections in $3.3\%$ of frames at a $0.35$ confidence threshold ($10.7\%$ at $0.25$), which the integrity gates filter out.

\emph{Integrity gating.} Our perception pipeline prioritizes false negatives over false positives: missing updates cause momentary inertial coasting, whereas incorrect position estimates can destabilize flight control. On the deployed gripper camera, three gates safeguard estimation: cluster support, kinematic feasibility ($0.8$\,m/s), and size consistency. On hovering recordings, most discards stem from missed detections ($75.5\%$) or incomplete depth ($18.4\%$), with corrupt background reads and size mismatches accounting for the remainder. Together, these gates ensure that $99.4\%$ of published live positions stay within $300$\,mm of truth.

\emph{Comparison to the fiducial baseline.} To benchmark against fiducial tracking, we evaluated both pipelines on the stereo camera setup of our earlier work~\cite{loritemora_clawar_2026} (Fig.~\ref{fig:accuracy}). On test figure-8 flights at $1.6$ to $2.6$\,m, the markerless stereo pipeline achieved a $122$\,mm median error. A rigid Horn-Markley alignment reveals that $106$\,mm was a constant reference-chain offset, isolating the markerless sensor precision at $71$\,mm compared to $12$ to $16$\,mm for the fiducial baseline. While fiducial markers offer higher raw precision, our markerless pipeline achieves sufficient accuracy for flight while eliminating physical markers.

\emph{Why the bare airframe needs the gates.} The stereo camera also highlights the depth challenge on nano-UAVs. Raw depth returns on the bare airframe yield a median of zero valid depth pixels out to $3.1$\,m. While affixing a $1.8$\,g matte panel restores depth returns, flight tests show the drone cannot support this payload: motor duty rises from $53$ to $55\%$ on the bare airframe to $80.5\%$ with the panel, causing motor saturation and emergency landing after $23$\,s. Gating protects against missing depth without adding airborne mass.

\IfFileExists{figures/gripper_accuracy_vs_distance.pdf}{%
\begin{figure}[!t]
\centering
\includegraphics[width=0.76\columnwidth]{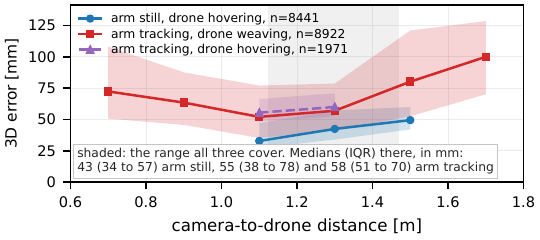}
\caption{Live accuracy of the deployed estimator (gripper camera) against motion capture over $17\,363$ positions from eight closed-loop flights. Each point is the median 3D error of the positions recorded within a $0.2$\,m interval of distance, the band around the curve their interquartile range, and intervals holding fewer than $80$ positions are not drawn. The gray span marks the distance range all three conditions cover, and the inset box gives each condition's median and interquartile range over that shared range.}
\label{fig:gripper}
\end{figure}}{}

\IfFileExists{figures/accuracy_vs_distance.pdf}{%
\begin{figure}[!t]
\centering
\includegraphics[width=0.76\columnwidth]{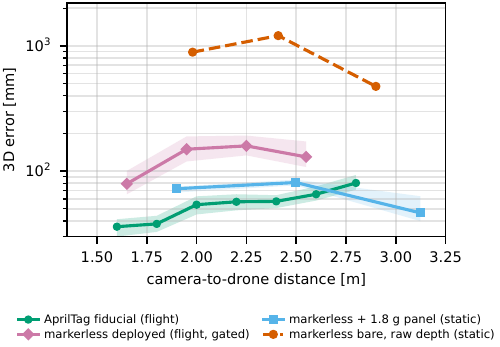}
\caption{Localization accuracy on the earlier stereo camera, comparing the fiducial baseline against three markerless configurations. Each point is the median 3D error of the positions recorded within a distance interval, and the band around a curve their interquartile range. The bare raw-depth series is medians only. Turning the gates on removes the bare airframe from the plot entirely, safely converting fatal false depths into dropouts ($4$ to $0$ wrong publications over $5\,539$ held-out frames).}
\label{fig:accuracy}
\label{tab:accuracy}
\label{fig:census}
\end{figure}}{}

\subsection{Active Observation on the Quadruped}
\label{sec:exp-tracking}
This experiment evaluates whether the perception-aware NMPC (Section~\ref{sec:pac}) improves drone framing compared to the reactive baseline. We quantify framing by the off-axis angle between the camera's optical axis and the drone, treating deviations above $30^\circ$ as tracking failures. During ten-waypoint flight tests, both controllers received motion-capture ground truth rather than markerless estimates, isolating controller tracking performance from perception noise (Table~\ref{tab:control}). The off-axis angle reconstructed offline from motion capture agrees with live controller telemetry to a median of $0.45^\circ$. The controller weight profiles, recorded flights, and evaluation scripts are provided in the open repository.

Table~\ref{tab:control} summarizes tracking performance. Across six reactive flights, the median off-axis angle remained between $10.5^\circ$ and $10.9^\circ$. The perception-aware NMPC substantially improved framing: the deployed tuning held a pooled median of $0.54^\circ$ (P95 $2.7^\circ$) over three flights, while a low-motion alternative maintained $0.70^\circ$. While all controllers kept the drone within the $30^\circ$ limit, NMPC provides critical margin during rapid maneuvers. The two NMPC tunings highlight an operational trade-off: allowing body rotation achieved a $0.70^\circ$ tracking angle while sweeping only $0.42\,\text{m}^2$ of floor area, whereas penalizing rotation commanded sidestepping to keep the background still, yielding $0.54^\circ$ tracking over a $1.72\,\text{m}^2$ footprint. We chose sidestepping because panning introduces novel background clutter that degrades detection (Section~\ref{sec:exp-closedloop}), while a stationary camera improves depth precision (Section~\ref{sec:exp-perception}). The reactive controller operated at a closer standoff ($0.86$\,m versus $1.25$\,m for the NMPC), which geometrically amplifies angular error for a given lateral displacement. Evaluating the reactive controller's lateral error at an equivalent $1.25$\,m standoff yields an off-axis angle of $7.4^\circ$, confirming that NMPC achieves a fourteen-fold improvement in framing precision.

\begin{table}[!t]
\caption{Perception-aware control on the quadruped.}
\label{tab:control}
\centering
\vspace{2pt}
\footnotesize
\setlength{\tabcolsep}{3pt}
\begin{tabular}{@{}lccccc@{}}
\hline
Controller & $n$ & Off-axis & Range & Body & Hand \\
 & & med / P95 ($^\circ$) & (m) & (m$^2$) & (m$^2$) \\
\hline
Reactive & 6 & 10.7 / 13.7 & 0.86 & 1.49 & 1.52 \\
NMPC, low-motion & 3 & 0.70 / 7.5 & 1.25 & 0.42 & 0.79 \\
NMPC, deployed & 3 & 0.54 / 2.7 & 1.25 & 1.72 & 1.61 \\
\hline
\end{tabular}
\par\vspace{3pt}
{\footnotesize\raggedright All rows are flown on the quadruped, motion-capture fed and scored, with no detector in the loop. Off-axis is the drone's angle off the camera axis (median / P95), range the median camera-to-drone distance, and body and hand the median convex-hull footprint swept per flight. Off-axis is reconstructed offline from motion capture, validated against the NMPC node's live telemetry (median residual $0.45^\circ$), and off-axis and range pool all airborne samples. The deployed tuning is the one flown in Sections~\ref{sec:exp-closedloop} and~\ref{sec:exp-ridethrough}.\par}
\end{table}

\IfFileExists{figures/exp3_combined.pdf}{%
\begin{figure*}[!t]
\centering
\includegraphics[width=\textwidth]{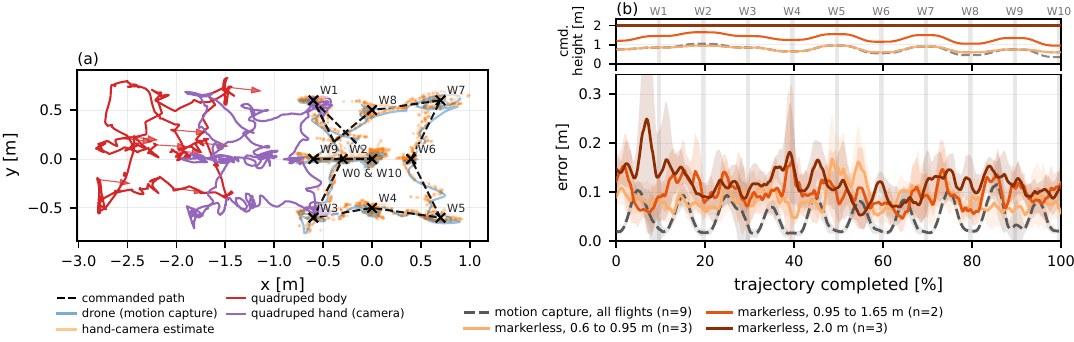}
\caption{The closed-loop flights of Section~\ref{sec:exp-closedloop}. (a)~One flight from above, with the commanded trajectory (dashed, waypoints W1 to W10) drawn over the translucent traces: the drone by motion capture (blue), the hand-camera estimate (orange), the quadruped's body heading (red), and the hand carrying the camera (purple). The loop begins and ends over the landing pad (W0 \& W10). (b)~Error against progress along the trajectory, defined as the instantaneous distance from the commanded position to the drone's true motion-capture position (mean $\pm$ one standard deviation). Shaded vertical bands indicate the $1.0$\,s hold at each waypoint. Gray traces represent the baseline control error from nine flights flown entirely on motion capture. Orange traces represent markerless flights, grouped by their commanded heights (top strip), which change dynamically at each waypoint.}
\label{fig:exp3_xy}
\end{figure*}}{}

\subsection{Closed-Loop Flight on Markerless Vision Alone}
\label{sec:exp-closedloop}
\label{sec:exp-usecases}
In this experiment the drone flies exclusively on the quadruped's markerless position stream from takeoff to landing. Motion capture records the flight strictly for evaluation and provides an external safety fallback. The drone's onboard EKF fuses IMU propagation with external position estimates. If no estimate arrives for $4.0$\,s, the bridge hands the drone back to motion capture as a safety fallback.

With the quadruped's arm held stationary, the drone completed three $120$\,s hovers at $1.3$\,m from the camera. Across these flights, the drone maintained hover station-keeping with a lateral error of $37$ to $53$\,mm (median $44$\,mm) relative to its setpoint, while the visual estimate differed from motion capture by a median offset of $35$ to $46$\,mm. Decomposed by axis, hover drift remained bounded within $24$\,mm laterally and $32$\,mm longitudinally, confirming that the onboard EKF does not oscillate under the vision stream. Next, with the active perception-aware NMPC tracking the drone, three flights executed a ten-waypoint trajectory, completing takeoff, navigation, and landing exclusively on the markerless estimates (Fig.~\ref{fig:exp3_xy}). Each flight lasted $124$\,s without a single hand-back to the safety system. Across all flights, visual estimates were available at $5$\,Hz or higher for the entire airborne duration, achieving a 3D median error of $58$ to $60$\,mm (P95 $121$ to $124$\,mm). The drone stayed within a median of $2.9^\circ$ of the optical axis throughout. To prevent background shifts from disrupting visual detection, the quadruped sidestepped along the room boundaries rather than rotating (Fig.~\ref{fig:exp3_xy}a, red trace).

Fig.~\ref{fig:exp3_xy}b compares trajectory tracking error across position sources. Error is evaluated against the moving setpoint rather than the static target waypoint. Flown on motion capture, aerodynamic tracking lag was $33$\,mm median. Under markerless vision, tracking error increased to $78$\,mm for low trajectories ($0.6$ to $0.95$\,m height), $88$\,mm for medium trajectories ($0.95$ to $1.65$\,m), and $109$\,mm for high trajectories ($2.0$\,m). Decomposing tracking error shows that residuals peak during direction reversals at waypoints W3 and W7, settling to within $45$\,mm along straight legs. This roughly two- to threefold increase is driven by $0.18$\,s pipeline latency, visual estimation noise ($56$\,mm median error), and the larger standoff required at higher elevations, which degrades depth resolution (Fig.~\ref{fig:gripper}). The drone tracks as tightly as latency permits, indicating that higher precision requires faster visual estimation rather than stiffer flight control.

\begin{table}[!t]
\caption{Flown envelope on a clear scene.}
\label{tab:envelope}
\centering
\scriptsize
\setlength{\tabcolsep}{1.6pt}
\begin{tabular}{@{}lclcccccc@{}}
\toprule
Drone height & $n$ & Config. & Range & Look-up & Avail. & \multicolumn{2}{c}{3D error (mm)} & Reached \\
(m) & & & (m) & med ($^\circ$) & (\%) & med & P95 & \\
\midrule
0.6 to 0.95 & 3 & S & 1.2 & 2 to 4 & 99.3 to 99.8 & 58 to 60 & 121 to 124 & 10/10 \\
0.95 to 1.65 & 2 & C & 1.2 & 31 to 33 & 98.5 to 99.6 & 50 to 59 & 131 to 143 & 10/10 \\
2.0 & 1 & C & 1.3 & 58 & 63.4 & 84 & 146 & 9/10 \\
2.0 & 2 & L & 1.8 & 44 to 45 & 77.7 to 84.5 & 87 to 89 & 150 to 153 & 10/10 \\
2.5 & 2 & C & {-} & {-} & {-} & {-} & {-} & 0/10 \\
\bottomrule
\end{tabular}
\par\vspace{2pt}
{\footnotesize\raggedright Each row groups flights at a given commanded height. Reached counts completed over commanded waypoints. All configurations fly the deployed NMPC controller (Table~\ref{tab:control}). S is the baseline configuration. C tightens the size-consistency gate to $1.5$. L increases the standoff to $1.8$\,m and frees wrist pitch to track higher elevations. Availability is airborne time outside estimate gaps longer than $0.15$\,s. No safety hand-backs occurred. Both $2.5$\,m attempts ended in the onboard watchdog landing during ascent.\par}
\end{table}

\IfFileExists{figures/occlusion_drift_axes.pdf}{%
\begin{figure}[!t]
\centering
\includegraphics[width=0.95\columnwidth]{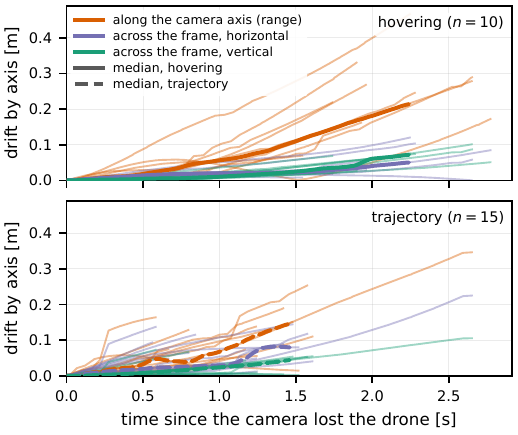}
\caption{Where a blind drone's drift goes. Each trace is one of $25$ human occlusions, separated by whether the drone was hovering (top, solid medians) or flying the trajectory (bottom, dashed medians). Drift is resolved on the camera's axes at the instant of loss: motion along the optical axis affects range, cross-frame motion affects tracking angle. The along-axis median reaches $0.22$\,m while cross-frame drift stays under $0.08$\,m. Medians terminate when fewer than five occlusions remain active.}
\label{fig:driftaxes}
\end{figure}

\begin{figure}[!t]
\centering
\includegraphics[width=0.95\columnwidth]{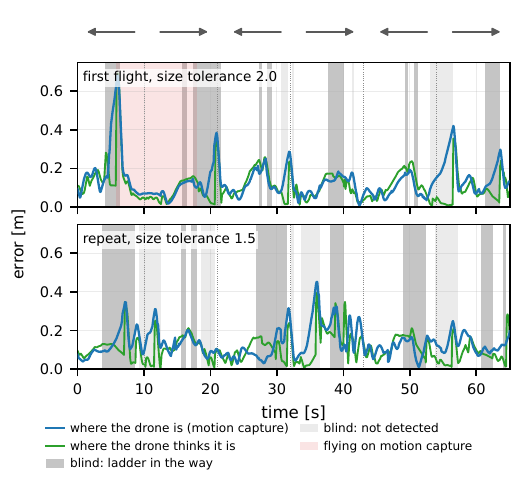}
\caption{Two flights behind a step ladder, against time from the first waypoint command, with top arrows indicating the direction of travel for each leg. Error is measured against motion capture (blue) and the drone's onboard filter (green), and the gap between them is the unobservable drift while blind. Shaded regions mark when the camera lost the drone: dark where a ladder rail crossed the line of sight, light where the line stayed clear and the detector instead failed against background clutter. Red marks the safety hand-back to the reference feed.}
\label{fig:ladder}
\end{figure}}{}

\subsection{Loss of Line of Sight and Flown Envelope}
\label{sec:exp-ridethrough}
Ten additional flights stress-tested the pipeline against visual occlusions and kinematic limits: a person crossing between the robots, a step ladder in the near field, and flights near the arm's elevation limit. During dropouts, the bridge sent no position updates, forcing the drone to coast on its IMU under two hierarchical timers. The first is an experimental safety switch that returns control to motion capture if no visual estimate arrives for $4.0$\,s. The second is an onboard failsafe watchdog that commands an emergency landing if all position streams cease for $4.5$\,s.

\textbf{Occlusion.} When a person walked between the robots, the camera experienced $36$ sustained occlusions lasting longer than $0.5$\,s, recovering visual tracking every time. The longest occlusion lasted $2.78$\,s. During these blind intervals, the drone drifted at approximately $0.14$\,m/s, accumulating up to $0.50$\,m of true physical drift while its internal IMU filter registered only $0.08$\,m. Inertial coasting succeeded because of the drift direction (Fig.~\ref{fig:driftaxes}): $64\%$ of the drift occurred longitudinally along the camera's optical axis. Because longitudinal motion changes range rather than pixel location, the drone shifted by a median of only $2.8^\circ$ off the optical axis during blind periods. It remained centered within the camera's $72^\circ$ diagonal field of view, enabling immediate visual re-acquisition once line of sight cleared. Across all occlusions, the camera regained track on the very first sample after line-of-sight recovery with a median error of $66$\,mm ($93$\,mm P90), correcting the accumulated blind drift by a median belief update of $0.16$\,m without triggering search maneuvers. No hand-back to motion capture was triggered during any of the $36$ occlusions, as the $2.78$\,s maximum blind duration remained below the $4.0$\,s safety threshold.

\textbf{Near-Field Clutter.} The drone flew behind a step ladder placed $0.6$\,m from the camera (Fig.~\ref{fig:ladder}). An initial flight with a permissive dimensionless size-consistency threshold of $2.0$ (the ratio of observed mask width to width predicted from depth) caused the estimator to lock onto the ladder rail's depth, pulling the drone $0.68$\,m off course and triggering a safety hand-back. Tightening this dimensionless threshold to $1.5$ rejected these spurious foreground depths. Across all $30\,428$ airborne frames, sweeping the threshold confirmed that $1.5$ is an optimal operating point, rejecting $48\%$ of false foreground depth locks ($43$ of $89$) while sacrificing only $0.35\%$ of valid frames. In a repeat flight with the tightened threshold, the drone completed its full trajectory autonomously, remaining blind for $48\%$ of the flight ($31.5$ of $65.0$\,s) but deviating at most $0.45$\,m from its planned path.

\textbf{Drone Flight Height and Viewing Elevation.} Table~\ref{tab:envelope} summarizes the operational tracking envelope. The system tracked dynamic trajectories up to $1.65$\,m above the floor with over $98\%$ estimate availability. At a flight height of $2.0$\,m above the floor and a $1.3$\,m standoff, the required camera pitch reached a median of $58^\circ$ and peaked at $82^\circ$, exceeding the arm's kinematic elevation limit of $68^\circ$ and reducing availability to $63.4\%$. Increasing the commanded standoff to $1.8$\,m reduced the required elevation to $45^\circ$, restoring availability to $77.7$ to $84.5\%$ and completing all waypoints. Attempting flight at $2.5$\,m height exceeded the arm's kinematic reach during the initial ascent, triggering the onboard failsafe watchdog landing. Across all flights, the camera tracked the drone over an $83^\circ$ elevation span, while azimuth variation remained within $43^\circ$ because the quadruped sidesteps rather than yawing.

\section{Discussion and Conclusion}

This work demonstrates that a quadruped robot can provide a closed-loop, markerless position stream accurate enough to pilot a nano-UAV, eliminating the need for payload-intensive onboard fiducials or external motion-capture systems. By combining a segmentation mask with registered depth and enforcing depth-integrity gating, the pipeline localized a bare nano-UAV using the quadruped's gripper camera with a 3D median error of $49$\,mm over $1\,069$\,s of closed-loop flight. Notably, the underlying detector was trained entirely on automatically generated masks, requiring zero manual labeling. Relying exclusively on this markerless stream, the drone executed extended $120$\,s hovers and autonomous ten-waypoint trajectories from takeoff to landing with a $58$ to $60$\,mm median localization error against motion capture. Concurrently, the quadruped used perception-aware NMPC to actively maintain the drone within the camera's field of view, while the drone's IMU safely coasted through transient line-of-sight occlusions lasting up to $2.8$\,s.

The primary generalizable insight of this work is the emphasis on outlier rejection over naive estimate smoothing. On lightweight airframes, depth-channel failures are frequent and severe. Our architecture demonstrates that converting spurious readings into dropouts is substantially safer than passing corrupted positions to the drone: an onboard IMU filter readily coasts through brief dropouts, whereas a confidently false position estimate destabilizes the flight controller. Evaluated on the stereo camera setup, these gates eliminated all meter-scale background locks across $5\,539$ test frames. A secondary insight is the value of establishing an empirical performance budget: our baseline confirmed that stable flight requires at least a $5$\,Hz update rate, which our pipeline comfortably exceeded by providing $20$\,Hz updates whenever the drone remained in view.

A central practical insight is the benefit of exploiting the quadruped's integrated hand camera rather than adding an external sensor payload. Commercial quadruped arms and grippers already package high-resolution optical and time-of-flight depth sensors accompanied by factory kinematic calibration. Repurposing these built-in sensors avoids mounting extra cameras, custom brackets, or auxiliary wiring harnesses on the quadruped, while the drone flies at its stock $37$\,g with no added payload. This demonstrates that collaborative heterogeneous localization requires no specialized sensory retrofitting when platforms share existing perception.

Several limitations bound our results. First, the detector was trained on laboratory recordings with uniform backgrounds, confining experiments to that environment. Rotating the camera introduced novel background clutter absent from training, requiring the quadruped to sidestep to maintain background consistency (Section~\ref{sec:exp-closedloop}). The drone's small, dark airframe returns sparse time-of-flight depth points, making depth extraction sensitive to surface orientation and ambient illumination. Second, the operational envelope is limited by the kinematic coupling between camera pitch and standoff distance. To maintain a stationary background for the detector, the quadruped keeps its camera level and tracks the drone primarily by adjusting arm height and standoff distance rather than pitching upward. Observing a higher drone forces the quadruped to increase standoff distance, pushing the target beyond the $2.3$\,m maximum effective range of the time-of-flight depth sensor. Tracking remains reliable between ranges of $0.6$ to $2.3$\,m and up to $1.65$\,m flight height, and degrades at $2.0$\,m (Section~\ref{sec:exp-ridethrough}). We expect a ceiling near $3.5$\,m by relaxing wrist pitch penalties for steeper camera tilt, though our laboratory ceiling bounded validation. Third, the architecture requires direct line-of-sight: once the drone navigates deep into cluttered structures, a single viewpoint loses observability~\cite{zhang_degeneracy_2016}. Finally, flight campaigns spanning three hovers, three full-loop autonomous flights, ten occlusion and clutter probe flights, twelve controller flights, and nine reference flights demonstrate practical viability rather than exhaustive statistical certifications.

Looking ahead, future work will focus on expanding robustness and generalizability toward real-world industrial deployments. First, visual limitations can be mitigated by training on diverse datasets with extensive augmentation, paired with airframe modifications such as high-contrast coloring and small matte surfaces for reliable depth extraction. Second, we plan to integrate cooperative path planning that coordinates joint robot trajectories over extended inspection missions. Finally, to overcome single-viewpoint observability limits in cluttered structures, future architectures could incorporate complementary modalities such as radio-frequency ranging, optical beacons, or onboard relative tracking.

\bibliographystyle{IEEEtran}
\bibliography{references}

\end{document}